\pdfoutput=1

\documentclass[11pt]{article}

\usepackage[preprint]{acl}

\usepackage{times}
\usepackage{latexsym}

\usepackage[T1]{fontenc}

\usepackage[utf8]{inputenc}

\usepackage{microtype}

\usepackage{inconsolata}

\usepackage{amssymb}
\usepackage{booktabs}
\usepackage{amsmath}
\usepackage{graphicx}

\title{Consistency Training by Synthetic Question Generation for Conversational Question Answering}

\author{Hamed Hematian Hemati \quad Hamid Beigy\\ \\
  AI Group, Computer Engineering Department, Sharif University of Technology}

\begin{document}
\maketitle

\begin{abstract}
Efficiently modeling historical information is a critical component in addressing user queries within a conversational question-answering (QA) context, as historical context plays a vital role in clarifying the user's questions. However, irrelevant history induces noise in the reasoning process, especially for those questions with a considerable historical context. In our novel model-agnostic approach, referred to as \textbf{CoTaH} (\textbf{Co}nsistency-\textbf{T}rained \textbf{a}ugmented \textbf{H}istory), we augment the historical information with synthetic questions and subsequently employ consistency training to train a model that utilizes both real and augmented historical data to implicitly make the reasoning robust to irrelevant history. To the best of our knowledge, this is the first instance of research using question generation as a form of data augmentation to model conversational QA settings. By citing a common modeling error prevalent in previous research, we introduce a new baseline model and compare our model's performance against it, demonstrating an improvement in results, particularly when dealing with questions that include a substantial amount of historical context. The source code can be found on our GitHub page.\footnote{\url{https://github.com/HamedHematian/SynCQG}}
\end{abstract}

\section{Introduction}
Humans often seek data through an information-seeking process, in which users engage in multiple interactions with machines to acquire information about a particular concept. Prominent examples of this phenomenon include the introduction of ChatGPT \cite{DBLP:journals/corr/abs-2303-08774} and the adoption of industrial systems like Amazon Alexa. Conversational Question-Answering (CQA) systems address user questions within the context of information-seeking interactions. In CQA, unlike conventional question answering, questions are interconnected, relying on previous questions and their corresponding answers to be fully understood without ambiguities. Although many researchers have proposed solutions to model history in CQA, a common modeling mistake made in these studies is using the gold answers of the history instead of the predicted ones. Our work aligns with the framework of addressing irrelevant history, as introduced by \citet{DBLP:conf/aaai/QiuHCJQ0HZ21}. However, unlike \citet{DBLP:conf/aaai/QiuHCJQ0HZ21}, our method abstains from utilizing the gold answers of history. Moreover, unlike \citet{DBLP:conf/aaai/QiuHCJQ0HZ21}, we utilize only one transformer during prediction, resulting in reduced time and memory. Initially, we augment the history of questions in the training set with synthetic questions. Our underlying idea is to maintain the model's consistency in its reasoning, whether utilizing the original historical data or the augmented version. Bert-HAE \cite{DBLP:conf/sigir/Qu0QCZI19} and HAM \cite{DBLP:conf/cikm/QuYQZCCI19} have previously served as baselines for several prior methods, but \citet{robust} conducted a re-implementation of these models using predicted history answers, which resulted in a significant performance decrease. As a result, in this paper, we employ the base transformer of our method as the baseline, as its performance surpasses the re-implementation of the mentioned methods. Our method results in a $1.8\%$ upgrade in overall F1 score, with causing a significant improvement in the scores of questions with a large historical context.

\section{Related Works}
The task of CQA has been introduced to extend question answering to a conversational setting. CoQA \cite{DBLP:journals/tacl/ReddyCM19} and QuAC \cite{DBLP:conf/emnlp/ChoiHIYYCLZ18} have been proposed as two extractive datasets in the CQA task. Bert-HAE \cite{DBLP:conf/sigir/Qu0QCZI19} employs a manually defined embedding layer to annotate tokens from previous answers within the document, and \citet{DBLP:conf/cikm/QuYQZCCI19} extends this approach introducing an ordering to these annotations. FlowQA \cite{DBLP:conf/iclr/HuangCY19} utilizes multiple blocks of Flow and Context Integration to facilitate the transfer of information between the context, the question, and the history. \citet{DBLP:conf/aaai/QiuHCJQ0HZ21} introduces the idea of irrelevant history and its effect in degrading performance, proposing a policy network to select the relevant history before reasoning. However, \citet{DBLP:conf/sigir/Qu0QCZI19, DBLP:conf/cikm/QuYQZCCI19, DBLP:conf/iclr/HuangCY19, DBLP:conf/aaai/QiuHCJQ0HZ21} employ the gold answers from history in their modeling. This approach deviates from real-world scenarios, where systems should rely on their previous predictions to answer current questions \cite{robust}. \citet{robust} re-implements Bert-HAE and HAM using the model's predictions, reporting a sharp decrease in performance. FlowQA experiences a performance drop from $64.4\%$ to $59.0\%$ on the development set when gold answers in history are not used \cite{DBLP:conf/iclr/HuangCY19}.

\section{Problem Definition}
To model a CQA setting, at dialog turn $k$, a model receives a question ($q_k$), a document containing the answer ($D$), and the history of the question ($H_k$), which is represented as a set of tuples, such as $H_k = \{(q_0, a_0^{pred}), \cdots, (q_{k-1}, a_{k-1}^{pred})\}$, where $a_i^{pred}$ is the model's prediction for $q_i$. It's important to note that the model may utilize only some of this information. For instance, we only employ history questions while excluding history answers. The primary objective is to predict the answer $a_k^{pred}$ for $q_k$.

\begin{equation}
    a_k^{pred} = \underset{a_k}{\arg\max} P(a_k|q_k,H_k,D)
\end{equation}

\section{Methodology} \label{method}
We seek to make the reasoning robust to irrelevant history implicitly by augmenting the dataset. To this end, for question $q_k$, we augment its history by injecting some synthetic questions. Let $H_k^{\star}$ be the augmented history. The intuition is that irrespective of whether the reasoning is performed with $H_k$ or with $H_k^{\star}$, the result should be the same. In other words:
\begin{equation}
    P(a_k|q_k, H_k, D) = P(a_k|q_k, H_k^{\star}, D)
    \label{imposed}
\end{equation}
To achieve this goal, we establish a two-stage pipeline. Our pipeline consists of a history augmentation module, whose goal is to augment the history and a question-answering module, whose objective is to consistently train a QA network so that the reasoning is consistent. The overall architecture of our model is depicted in Figure \ref{fig:atch}.

\subsection{History Augmentation Module}
This module includes a conversational question generator, denoted as $CQG_{\theta}$, where $\theta$ represents the parameter set of the generator, and a question selector, denoted as $QS$, which is responsible for choosing a set of $S$ synthetic questions generated to augment the history.

\paragraph{Training}
The first step involves training $CQG_\theta$. While there has been research aimed at generating conversational questions \cite{DBLP:conf/eacl/GuMYS21, DBLP:conf/acl/PanLYCS19}, for the sake of simplifying the implementation, we employ a straightforward generative transformer for this task. To train this network, we input $D$, $H_k$, and $a_k$ into the network, intending to generate $q_k$. We train this network using cross-entropy loss in an auto-regressive manner. In \ref{qg}, question generation result is described.

\begin{figure*}[h!]
	\begin{center}
		\includegraphics[width=1.\textwidth]{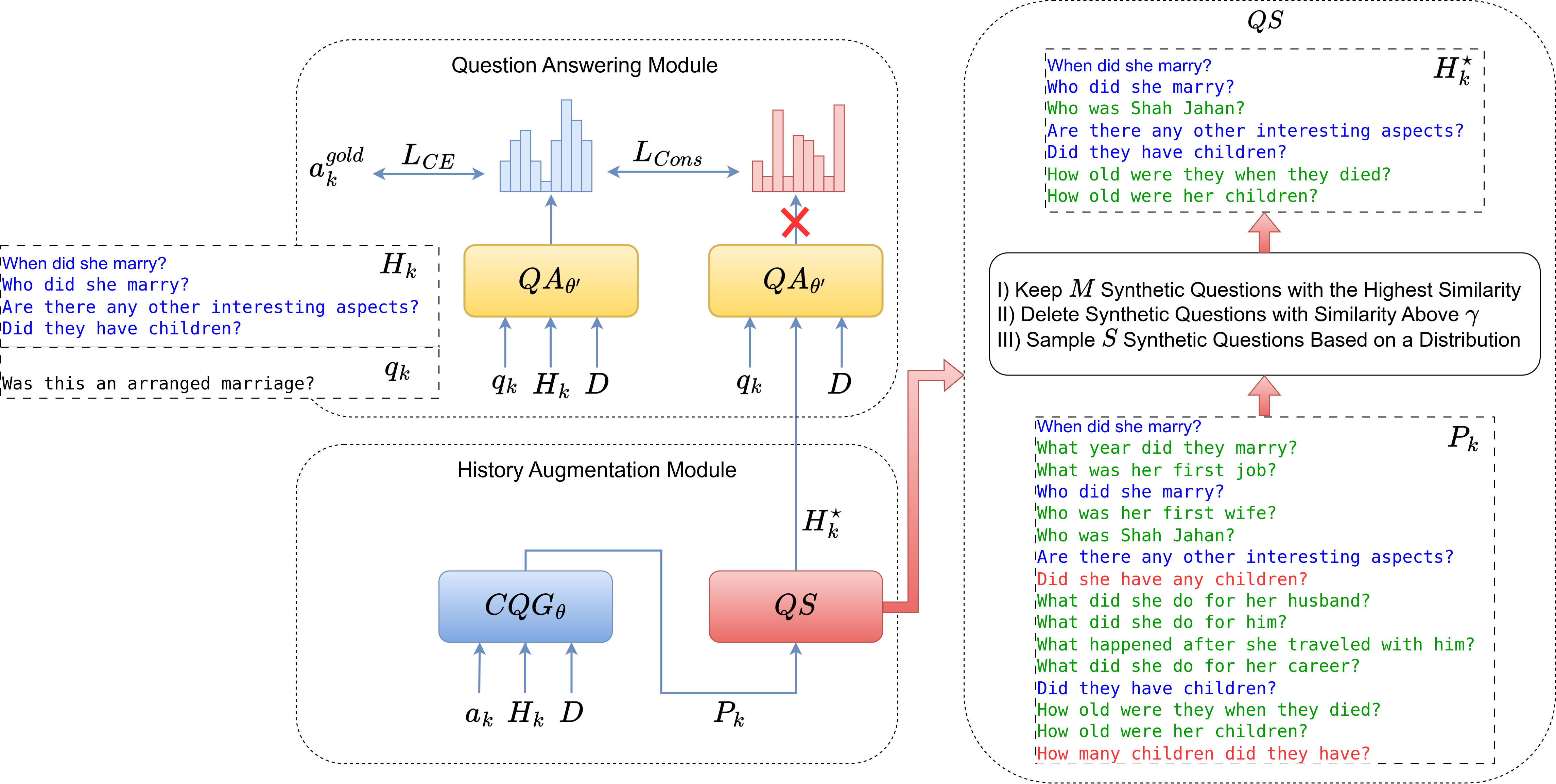}
		\caption
		{
            \textbf{Architecture of the Model}: For a given question $q_k$, the conversational question generator $CQG_\theta$ constructs a pool of questions denoted as $P_k$. questions in $H_k$ are shown in blue, and synthetic questions are depicted in green and red. The synthetic questions, which are similar to $H_k$ questions, are marked in red, while dissimilar ones are in green. The question selector $QS$ discards red synthetic questions, selects $M$ ones with the highest scores, and chooses $S=3$ synthetic questions from the green questions according to uniform distribution, along with $H_k$ questions, to create $H_k^{\star}$. The QA network $QA_{\theta'}$ computes its output using both $H_k$ and $H_k^{\star}$ as input. The QA network is trained by minimizing $L_{CE}$ and $L_{Cons}$.
		}
		\label{fig:atch}
	\end{center}
\end{figure*}

\paragraph{Question Generation}
After training $CQG_\theta$, we aim to generate synthetic conversational questions for the training set. Suppose that we want to generate synthetic conversational questions between $q_k$ and $q_{k+1}$. Suppose that $a_k$ is located in the $i$-th sentence of the document. We extract noun phrases from sentences $i-1$, $i$, and $i+1$ as potential answers. We make this choice as we want these answers to be similar to the flow of conversation and if these answers are extracted from local regions, the likelihood increases. Let one of these answers be called $a_k^{syn}$. We feed $D$, $H_k$, $a_k^{syn}$ to $CQG_{\theta}$ to obtain $q_k^{syn}$. We iterate this process for all $0 \leq j \leq k$, and generate synthetic questions. We refer to all generated synthetic questions and real questions of the history as pool of questions ($P_k$) for $q_k$.

\paragraph{Question Filtering \& Injection}
We could set $P_k$ as $H_k^{\star}$, however, $P_k$ contains a multitude of synthetic questions which induces too much noise. Additionally, in the consistency training setting, the noise (perturbation) should be small. Thus, we only select $S$ of synthetic questions from $P_k$, where $S$ is a hyperparameter. Not all synthetic questions are helpful, necessitating the need to filter out degenerate ones. We want our selected synthetic questions to be similar to the trend of the conversation. To this end, we compute a score for each synthetic question and only keep the top $M$ synthetic questions with the highest score. To compute the score, each question (real or synthetic) is encoded with LaBSE \cite{DBLP:conf/acl/FengYCA022}. For each synthetic question $q^{syn}$ which is located between history turns $q_i$ and $q_{i+1}$, the score is computed as $Sim(h(q_i), h(q^{syn})) + Sim(h(q_{i+1}), h(q^{syn}))$, where $Sim$ is the cosine similarity function and $h(\mathbf{x})$ is the LaBSE's encoding of the sentence $\mathbf{x}$. Additionally, Sometimes, we generate questions that are too similar to previous or future questions, which are invaluable. Thus, we compare the similarity of generated question $q^{syn}$ with questions in $\{q_k\} \bigcup H_k$ and if the similarity is above $\gamma$, $q^{syn}$ is discarded. This situation is depicted in Figure \ref{fig:atch}, where $P_k$ contains real history questions, depicted in blue, and synthetic questions, depicted in red and green. Those synthetic questions that have high similarity with $\{q_k\} \bigcup H_k$, are depicted in red. As it can be seen two questions ``Did she have any children'' and ``How many children did they have'' have high similarity with the question ``Did they have children'', and thus, they're discarded. The effectiveness of question filtering is approved in Section \ref{qf_res}. In addition, we need to set a distribution to guide the selection of $S$ number of generated questions, for which we adopt a uniform distribution. More details on the distribution selection are available in Section \ref{dist}.

\begin{table*}[htb] 
	\selectfont
	\begin{center}
		\caption{Comparison of our methods with other benchmarks on the test set}
		\label{table:table_comp}
		\begin{tabular}{c|cccc}
			\toprule
			Model Name
			& F1
			& HEQ-Q 
			& HEQ-D 
			& Unrealistic Settings\\
			\midrule
			Bert-HAE-Real \cite{robust}
			&  53.5  & -  & - & \\
			HAM-Real \cite{robust}
			&  54.2  &  - & - & \\	
			Bert (Our Model)
			&  58.9  & 52.9  &  5.3 & \\
			CoTaH-Bert (Our Model)
			&  60.7  & 55.3  & 5.9 & \\
			Bert-HAE \cite{DBLP:conf/sigir/Qu0QCZI19}
			&  62.4  & 57.8  & 5.1 & $\checkmark$\\
			HAM \cite{DBLP:conf/cikm/QuYQZCCI19}
			&  64.4  &  60.2 & 6.1 & $\checkmark$\\
			Reinforced Backtracking \cite{DBLP:conf/aaai/QiuHCJQ0HZ21}
			&  66.1  &  62.2 &  7.3 & $\checkmark$\\
			\bottomrule
		\end{tabular}
	\end{center}
\end{table*}

\subsection{Question Answering Module}
For each question $q_k$, as illustrated in Figure \ref{fig:atch}, we feed $q_k$, $H_k$, and $D$ to the QA network ($QA_{\theta'}$) to compute the answer distribution. In parallel, we feed $q_k$, $H_k^{\star}$, and $D$ to the QA network to compute another answer distribution. As mentioned in Section \ref{method}, we need to impose the condition outlined in Equation \ref{imposed}. To achieve this, we employ KL-Divergence between the answer distributions. Additionally, we use cross-entropy loss to train the QA network for answer prediction. The losses are calculated as per Equation \ref{eq:loss}, where $L_{CE}$, $L_{Cons}$, and $L_{T}$ represent the cross-entropy loss, consistency loss, and total loss. $\lambda$ is a hyperparameter used to determine the ratio of the two losses. 

\begin{align}
		&L_{CE}=CE(QA_{\theta'}(q_k,H_k,D),a_k^{gold}) \nonumber \\
		&L_{Cons}=D_{KL}(QA_{\theta'}(q_k,H_k,D), \label{eq:loss}\\
		&\:\:\:\:\:\:\:\:\:\:\:\:\:\:\:\:\:\:\:\:\:\:\:\:\:\:\:QA_{\theta'}(q_k,H_k^{\star},D)) \nonumber\\
		&L_{T}=L_{CE} + \lambda L_{Cons}  \nonumber
\end{align}

Furthermore, we acknowledge that augmenting the history for all questions may not be optimal, as initial questions in a dialog, due to their little historical context, may not require augmentation for robust reasoning. In this case augmenting their history might add unnecessary noise, potentially degrading performance. Thus, we introduce a threshold named $\tau$ and only augment the history of $q_k$ if $k \geq \tau$. According to \citet{DBLP:journals/pami/MiyatoMKI19}, we only pass the gradients through one network. As shown in the Figure \ref{fig:atch}, the symbol \textcolor{red}{$\times$} is used to denote gradient cut. It should be noted that our method is model-agnostic, and any architecture could be used as the QA network.

\section{Setup}
We utilize the QuAC dataset \cite{DBLP:conf/emnlp/ChoiHIYYCLZ18}, to conduct our experiments on, and data splitting is described in \ref{HG}.  We utilize Bert \cite{DBLP:conf/naacl/DevlinCLT19} as our base model to conduct experiments following the previous research. For question generation, we adopt Bart-Large \cite{DBLP:conf/acl/LewisLGGMLSZ20}. Following \citet{DBLP:conf/emnlp/ChoiHIYYCLZ18}, we use F1, HEQ-Q, and HEQ-D as our evaluation metrics. F1 measures the overlap between $a_k^{gold}$ and $a_k^{pred}$. HEQ-Q and HEQ-D are the ratio of questions and dialogs, for which the model performs better than human \cite{DBLP:conf/emnlp/ChoiHIYYCLZ18}. In Section \ref{hyp}, the process of choosing all other hyperparameters and their analysis is described. For all of our models, we concatenate the question with history questions, feeding them to the network. More details on reproducibility are presented in Section \ref{rep}.

\section{Results}
In Table \ref{table:table_comp}, we have depicted our results on the test set division in comparison to previous relevant models. It should be noted that our test set is different from previous methods, but it has been drawn from the same distribution. As stated before, Bert-HAE and HAM leverage the gold answers of history. Their re-implementations by \citet{robust} are shown in the Table as Bert-HAE-Real and HAM-Real, which indicate a significant drop in performance. In this scenario where common baselines experience a substantial decrease, we examine a basic Bert model with history concatenation as the baseline, as its performance is superior. Our model outperforms this baseline by $1.8\%$ in the F1 score. According to Figure \ref{fig:test_turns}, this improvement is mostly due to an improvement in the performance of questions with a large amount of history. This confirms that our intuition is valid that our method enhances the base model's ability to answer questions with a large historical context. Moreover, while Bert-HAE outperforms CoTaH-Bert in terms of F1 score, CoTaH-Bert exhibits superior performance in HEQ-D. This highlights the better consistency of our model to maintain its performance throughout the entire dialog, which is achieved through superiority in answering the questions in the latter turns. Additionally, we include the results of the history backtracking model \cite{DBLP:conf/aaai/QiuHCJQ0HZ21} in the table. Since this model's code is not publicly available, we have been unable to re-implement it with the correct settings and perform a meaningful comparison. However, it's worth noting that this model utilizes unrealistic settings in two stages: once for history selection and once for question answering, potentially exacerbating the modeling issues even further. We have used ``Unrealistic Settings'' as a term to indicate that a model uses gold answers of history in its modeling. 

\section{Conclusions}
In this paper, we introduced a novel model-agnostic method to make the reasoning of conversational question-answering models robust to irrelevant history. We coped with this issue by augmenting the history and training the model with consistency training. In our experiments, we didn't follow the wrong modeling of past research in using the gold answers of history. We examined our method with Bert which exhibited a $1.8\%$ performance boost compared to the baseline model. It was demonstrated that this improvement is primarily attributed to the enhancement of the model's performance on questions with a substantial historical context, suggesting that our method has been successful in making the reasoning robust for these questions.

\section{Limitations}
Our model requires a phase of question generation. For synthetic question generation, the history augmentation module could be slow and the speed is directly correlated to the number of questions that one opts to generate. However, question generation is trained only once and all questions are generated in a single run, and all of other experiments are conducted by only training the QA module. Moreover, although our model doesn't need any further computation during evaluation than merely running the QA network, we need two forward passes during the training phase, which makes the training of the QA network a bit more time-consuming than training the baseline model.

\bibliography{custom}

\begin{thebibliography}{17}
\expandafter\ifx\csname natexlab\endcsname\relax\def\natexlab#1{#1}\fi

\bibitem[{Choi et~al.(2018)Choi, He, Iyyer, Yatskar, Yih, Choi, Liang, and
  Zettlemoyer}]{DBLP:conf/emnlp/ChoiHIYYCLZ18}
Eunsol Choi, He~He, Mohit Iyyer, Mark Yatskar, Wen{-}tau Yih, Yejin Choi, Percy
  Liang, and Luke Zettlemoyer. 2018.
\newblock \href {https://doi.org/10.18653/v1/d18-1241} {Quac: Question
  answering in context}.
\newblock In \emph{Proceedings of the 2018 Conference on Empirical Methods in
  Natural Language Processing, Brussels, Belgium, October 31 - November 4,
  2018}, pages 2174--2184. Association for Computational Linguistics.

\bibitem[{Devlin et~al.(2019)Devlin, Chang, Lee, and
  Toutanova}]{DBLP:conf/naacl/DevlinCLT19}
Jacob Devlin, Ming{-}Wei Chang, Kenton Lee, and Kristina Toutanova. 2019.
\newblock \href {https://doi.org/10.18653/v1/n19-1423} {{BERT:} pre-training of
  deep bidirectional transformers for language understanding}.
\newblock In \emph{Proceedings of the 2019 Conference of the North American
  Chapter of the Association for Computational Linguistics: Human Language
  Technologies, {NAACL-HLT} 2019, Minneapolis, MN, USA, June 2-7, 2019, Volume
  1 (Long and Short Papers)}, pages 4171--4186. Association for Computational
  Linguistics.

\bibitem[{Feng et~al.(2022)Feng, Yang, Cer, Arivazhagan, and
  Wang}]{DBLP:conf/acl/FengYCA022}
Fangxiaoyu Feng, Yinfei Yang, Daniel Cer, Naveen Arivazhagan, and Wei Wang.
  2022.
\newblock \href {https://doi.org/10.18653/V1/2022.ACL-LONG.62}
  {Language-agnostic {BERT} sentence embedding}.
\newblock In \emph{Proceedings of the 60th Annual Meeting of the Association
  for Computational Linguistics (Volume 1: Long Papers), {ACL} 2022, Dublin,
  Ireland, May 22-27, 2022}, pages 878--891. Association for Computational
  Linguistics.

\bibitem[{Gu et~al.(2021)Gu, Mirshekari, Yu, and
  Sisto}]{DBLP:conf/eacl/GuMYS21}
Jing Gu, Mostafa Mirshekari, Zhou Yu, and Aaron Sisto. 2021.
\newblock \href {https://doi.org/10.18653/V1/2021.EACL-MAIN.177} {Chaincqg:
  Flow-aware conversational question generation}.
\newblock In \emph{Proceedings of the 16th Conference of the European Chapter
  of the Association for Computational Linguistics: Main Volume, {EACL} 2021,
  Online, April 19 - 23, 2021}, pages 2061--2070. Association for Computational
  Linguistics.

\bibitem[{Huang et~al.(2019)Huang, Choi, and Yih}]{DBLP:conf/iclr/HuangCY19}
Hsin{-}Yuan Huang, Eunsol Choi, and Wen{-}tau Yih. 2019.
\newblock \href {https://openreview.net/forum?id=ByftGnR9KX} {Flowqa: Grasping
  flow in history for conversational machine comprehension}.
\newblock In \emph{7th International Conference on Learning Representations,
  {ICLR} 2019, New Orleans, LA, USA, May 6-9, 2019}. OpenReview.net.

\bibitem[{Lewis et~al.(2020)Lewis, Liu, Goyal, Ghazvininejad, Mohamed, Levy,
  Stoyanov, and Zettlemoyer}]{DBLP:conf/acl/LewisLGGMLSZ20}
Mike Lewis, Yinhan Liu, Naman Goyal, Marjan Ghazvininejad, Abdelrahman Mohamed,
  Omer Levy, Veselin Stoyanov, and Luke Zettlemoyer. 2020.
\newblock \href {https://doi.org/10.18653/V1/2020.ACL-MAIN.703} {{BART:}
  denoising sequence-to-sequence pre-training for natural language generation,
  translation, and comprehension}.
\newblock In \emph{Proceedings of the 58th Annual Meeting of the Association
  for Computational Linguistics, {ACL} 2020, Online, July 5-10, 2020}, pages
  7871--7880. Association for Computational Linguistics.

\bibitem[{Lin(2004)}]{lin2004rouge}
Chin-Yew Lin. 2004.
\newblock Rouge: A package for automatic evaluation of summaries.
\newblock In \emph{Text summarization branches out}, pages 74--81.

\bibitem[{Miyato et~al.(2019)Miyato, Maeda, Koyama, and
  Ishii}]{DBLP:journals/pami/MiyatoMKI19}
Takeru Miyato, Shin{-}ichi Maeda, Masanori Koyama, and Shin Ishii. 2019.
\newblock \href {https://doi.org/10.1109/TPAMI.2018.2858821} {Virtual
  adversarial training: {A} regularization method for supervised and
  semi-supervised learning}.
\newblock \emph{{IEEE} Trans. Pattern Anal. Mach. Intell.}, 41(8):1979--1993.

\bibitem[{OpenAI(2023)}]{DBLP:journals/corr/abs-2303-08774}
OpenAI. 2023.
\newblock \href {https://doi.org/10.48550/ARXIV.2303.08774} {{GPT-4} technical
  report}.
\newblock \emph{CoRR}, abs/2303.08774.

\bibitem[{Pan et~al.(2019)Pan, Li, Yao, Cai, and Sun}]{DBLP:conf/acl/PanLYCS19}
Boyuan Pan, Hao Li, Ziyu Yao, Deng Cai, and Huan Sun. 2019.
\newblock \href {https://doi.org/10.18653/V1/P19-1203} {Reinforced dynamic
  reasoning for conversational question generation}.
\newblock In \emph{Proceedings of the 57th Conference of the Association for
  Computational Linguistics, {ACL} 2019, Florence, Italy, July 28- August 2,
  2019, Volume 1: Long Papers}, pages 2114--2124. Association for Computational
  Linguistics.

\bibitem[{Papineni et~al.(2002)Papineni, Roukos, Ward, and
  Zhu}]{DBLP:conf/acl/PapineniRWZ02}
Kishore Papineni, Salim Roukos, Todd Ward, and Wei{-}Jing Zhu. 2002.
\newblock \href {https://doi.org/10.3115/1073083.1073135} {Bleu: a method for
  automatic evaluation of machine translation}.
\newblock In \emph{Proceedings of the 40th Annual Meeting of the Association
  for Computational Linguistics, July 6-12, 2002, Philadelphia, PA, {USA}},
  pages 311--318. {ACL}.

\bibitem[{Qiu et~al.(2021)Qiu, Huang, Chen, Ji, Qu, Wei, Huang, and
  Zhang}]{DBLP:conf/aaai/QiuHCJQ0HZ21}
Minghui Qiu, Xinjing Huang, Cen Chen, Feng Ji, Chen Qu, Wei Wei, Jun Huang, and
  Yin Zhang. 2021.
\newblock \href {https://doi.org/10.1609/AAAI.V35I15.17617} {Reinforced history
  backtracking for conversational question answering}.
\newblock In \emph{Thirty-Fifth {AAAI} Conference on Artificial Intelligence,
  {AAAI} 2021, Thirty-Third Conference on Innovative Applications of Artificial
  Intelligence, {IAAI} 2021, The Eleventh Symposium on Educational Advances in
  Artificial Intelligence, {EAAI} 2021, Virtual Event, February 2-9, 2021},
  pages 13718--13726. {AAAI} Press.

\bibitem[{Qu et~al.(2019{\natexlab{a}})Qu, Yang, Qiu, Croft, Zhang, and
  Iyyer}]{DBLP:conf/sigir/Qu0QCZI19}
Chen Qu, Liu Yang, Minghui Qiu, W.~Bruce Croft, Yongfeng Zhang, and Mohit
  Iyyer. 2019{\natexlab{a}}.
\newblock \href {https://doi.org/10.1145/3331184.3331341} {{BERT} with history
  answer embedding for conversational question answering}.
\newblock In \emph{Proceedings of the 42nd International {ACM} {SIGIR}
  Conference on Research and Development in Information Retrieval, {SIGIR}
  2019, Paris, France, July 21-25, 2019}, pages 1133--1136. {ACM}.

\bibitem[{Qu et~al.(2019{\natexlab{b}})Qu, Yang, Qiu, Zhang, Chen, Croft, and
  Iyyer}]{DBLP:conf/cikm/QuYQZCCI19}
Chen Qu, Liu Yang, Minghui Qiu, Yongfeng Zhang, Cen Chen, W.~Bruce Croft, and
  Mohit Iyyer. 2019{\natexlab{b}}.
\newblock \href {https://doi.org/10.1145/3357384.3357905} {Attentive history
  selection for conversational question answering}.
\newblock In \emph{Proceedings of the 28th {ACM} International Conference on
  Information and Knowledge Management, {CIKM} 2019, Beijing, China, November
  3-7, 2019}, pages 1391--1400. {ACM}.

\bibitem[{Reddy et~al.(2019)Reddy, Chen, and
  Manning}]{DBLP:journals/tacl/ReddyCM19}
Siva Reddy, Danqi Chen, and Christopher~D. Manning. 2019.
\newblock \href {https://doi.org/10.1162/tacl\_a\_00266} {Coqa: {A}
  conversational question answering challenge}.
\newblock \emph{Trans. Assoc. Comput. Linguistics}, 7:249--266.

\bibitem[{Siblini et~al.(2021)Siblini, Sayil, and Kessaci}]{robust}
Wissam Siblini, Baris Sayil, and Yacine Kessaci. 2021.
\newblock \href {https://doi.org/10.18653/v1/2021.acl-short.130} {Towards a
  more robust evaluation for conversational question answering}.
\newblock In \emph{Proceedings of the 59th Annual Meeting of the Association
  for Computational Linguistics and the 11th International Joint Conference on
  Natural Language Processing, {ACL/IJCNLP}, Virtual Event}, pages 1028--1034.

\bibitem[{Zhang et~al.(2020)Zhang, Kishore, Wu, Weinberger, and
  Artzi}]{DBLP:conf/iclr/ZhangKWWA20}
Tianyi Zhang, Varsha Kishore, Felix Wu, Kilian~Q. Weinberger, and Yoav Artzi.
  2020.
\newblock \href {https://openreview.net/forum?id=SkeHuCVFDr} {Bertscore:
  Evaluating text generation with {BERT}}.
\newblock In \emph{8th International Conference on Learning Representations,
  {ICLR} 2020, Addis Ababa, Ethiopia, April 26-30, 2020}. OpenReview.net.

\end{thebibliography}

\section{Appendix}

\subsection{Data Splitting} \label{HG}
Since the test set of QuAC is not publicly available, we divide the development (dev) set into dev/test sets randomly, such that the number of questions in dev and test sets is almost equal. The total number of dev and test questions is 3678 and 3676 respectively after splitting. In our splitting, each dialog, with all of its questions, is either attributed to the dev set or the test set, in order to prevent test data leakage. Further, according to \citet{DBLP:conf/emnlp/ChoiHIYYCLZ18}, original dev set of QuAC contains unique documents, meaning that a single document will not be shared among the final dev and test sets, potentially preventing test data leakage.

\subsection{Question Generation Results} \label{qg}
The results of question generation are evaluated in Table \ref{table:qg}. These scores are obtained from the dev data. Bleu-1,4 \cite{DBLP:conf/acl/PapineniRWZ02}, Rouge \cite{lin2004rouge}, and Bert-Score \cite{DBLP:conf/iclr/ZhangKWWA20} are used for criteria. We use the evaluate library\footnote{https://github.com/huggingface/evaluate} to implement these metrics. \citet{DBLP:conf/eacl/GuMYS21} reports better results for the question generation, yet we didn't aim to optimize Bart-Large meticulously as the generated questions have a good quality for our task. The point is that in this research, we only utilize questions alone without considering answers. Thus, if the generated questions have less correlations with answers, it's tolerable as they are still relevant questions considering the overall flow of conversation. it should be noted that if a future research wants to incorporate predicted answers in its modeling, it should be more cautious about the quality of the question generation to ensure that the right synthetic questions are generated concerning their answers.

\begin{table}[htb] 
    \caption{Question generation results on the dev set}
    \label{table:qg}
	\selectfont
	\begin{center}
		\begin{tabular}{cccc}
			\toprule
			Bleu-1 & Bleu-4 & Rouge-L & Bert-Score \\
			\midrule
			33.6 & 9.5 & 29.0 & 90.5 \\
			\bottomrule
		\end{tabular}
	\end{center}
\end{table}

\subsection{Hyperparameter Selection \& Sensitivity Analysis} \label{hyp}
Initially, we determine $M$ and $\gamma$ by assessing some examples of the dev data, setting $M=10$ and $\gamma=0.8$ based on our appraisal. Next, we determine the values of $S$, $\lambda$, and $\tau$ by conducting experiments on the dev set. In Table \ref{table:hyp}, we evaluate the effects of the model's two main hyperparameters, $S$ and $\lambda$, through a grid search with the following values: $S \in \{ 1,2,3 \}$ and $\lambda \in \{ 1.0,1.5,2.0 \}$. Firstly, it is evident that the model performs better when $S \in \{ 1,2 \}$ compared to when $S=3$ overall. This suggests that $S=3$ introduces too much noise, which could be detrimental for performance. Furthermore, when $\lambda \in { 1.5, 2.0 }$, the performance is better compared to $\lambda=1.0$, indicating that the introduction of $\lambda$ is helpful, as simply adding $L_{CE}$ and $L_{KL}$ (or equally setting $\lambda=1.0$) produces inferior performance. For the remaining experiments, we set $S=2$ and $\lambda=2.0$ as these settings yield the best F1 and HEQ-Q scores.

\begin{table}[htb]
\centering
\caption{The effect of $S$ and $\lambda$ on the dev set}
\label{table:hyp}
\begin{tabular}{c|c|ccc} \toprule
&  & \multicolumn{1}{c}{F1} & \multicolumn{1}{c}{HEQ-Q} & \multicolumn{1}{c}{HEQ-D} \\ 
  \midrule
    &$\lambda=1.0$ & 58.6 & 53.5 & 4.8 \\
    \textbf{$S=1$} &$\lambda=1.5$ & 59.1 & 54.8 & \textbf{5.5} \\
    &$\lambda=2.0$ & 59.0 & 54.2 & 4.4  \\
  \midrule
    &$\lambda=1.0$ &  57.9 & 52.7 & 4.0 \\
    \textbf{$S=2$} &$\lambda=1.5$ &  58.2 & 53.5 & 4.2  \\
    &$\lambda=2.0$ &  \textbf{59.4} &\textbf{54.8} & 5.1  \\
  \midrule
    &$\lambda=1.0$ & 58.3 & 53.5 & 5.1  \\
    \textbf{$S=3$} &$\lambda=1.5$ & 58.6 & 53.5 & 5.0 \\
    &$\lambda=2.0$ &  58.8 & 54.1 & 4.2 \\
  \bottomrule
\end{tabular}
\end{table}

After setting the right amount for $S$ and $\lambda$, we opt to examine whether the introduction of $\tau$ is effective. Thus, we conduct experiments on three different amount of this hyperparameter. In Table \ref{table:tau}, it's evident that the right amount of $\tau$ has a considerable effect on the performance, confirming our intuition about the functionality of $\tau$. For all tested values of $\tau$ within the set $\{5, 6, 7\}$, performance has increased compared to the base settings with $\tau=0$ (or equivalently, using no threshold). Notably, the maximum performance improvement is observed when $\tau=6$.

\begin{table}[htb]
\centering
\caption{The effect of $\tau$ on the dev set}
\label{table:tau}
\begin{tabular}{c|cccc} \toprule
 & \multicolumn{1}{c}{F1} & \multicolumn{1}{c}{HEQ-Q} & \multicolumn{1}{c}{HEQ-D} \\ 
  \midrule
    $\tau=0$ &  59.4 & 54.8 & 5.1 \\
    $\tau=5$ &  59.6 & 55.2 & 5.5  \\
    $\tau=6$ &  \textbf{59.9} & \textbf{55.2} & \textbf{5.5}  \\
    $\tau=7$ &  59.5 & 54.9 & 5.1 \\
  \bottomrule
\end{tabular}
\end{table}

\subsection{Question Filtering Effect} \label{qf_res}
After determining the optimal $\tau$, the effectiveness of the question-filtering, as discussed earlier, is examined. The results in Table \ref{table:filter} demonstrate that this filtering leads to a considerable additional performance boost by filtering out degenerate questions.

\begin{table}[htb]
\centering
\caption{The effect of question filtering on the dev set}
\label{table:filter}
\begin{tabular}{cccc} \toprule
Filtering Type& \multicolumn{1}{c}{F1} & \multicolumn{1}{c}{HEQ-Q} & \multicolumn{1}{c}{HEQ-D} \\ 
\midrule
    No Filtering & 59.9 & 55.2 & \textbf{5.5}  \\
    Similarity Filtering &  \textbf{60.9} & \textbf{56.3} & 5.3  \\
\bottomrule
\end{tabular}
\end{table}

\subsection{Synthetic Question Selection Distribution} \label{dist}
Although we select synthetic questions using a uniform distribution, we have conducted experiments using two distributions: uniform and linear. In the uniform setting, the generated questions are selected with the same probability. For the linear, if $q^{syn}$ is located between $q_j$ and $q_{j+1}$, its probability of being selected ($P(q^{syn})$) is $P(q^{syn}) \propto \: k-j$. We opt for the linear distribution, as we believe that closer synthetic questions to the original question might contribute to greater robustness, as questions that are further away are likely less relevant. The results are shown in Table \ref{table:q_sampling}. We observe a relatively $1\%$ drop in both F1 and HEQ-Q scores with the linear distribution, concluding that our hypothesis has not been true. Given the superiority of the uniform distribution, we choose to continue with it.

\begin{table}[h]
\centering
\caption{The effect of question selection distribution on the dev set}
\label{table:q_sampling}
\begin{tabular}{cccc} \toprule
Q-Selection Dist. & \multicolumn{1}{c}{F1} & \multicolumn{1}{c}{HEQ-Q} & \multicolumn{1}{c}{HEQ-D} \\ 
  \midrule
    Uniform & \textbf{60.9} & \textbf{56.3} & 5.3 \\
    Linear & 59.9 & 55.2 & \textbf{5.9} \\
\bottomrule
\end{tabular}
\end{table}

\subsection{Additional Results} \label{res}
\begin{figure}[h!]
 	\begin{center}
 		\includegraphics[width=.45\textwidth]{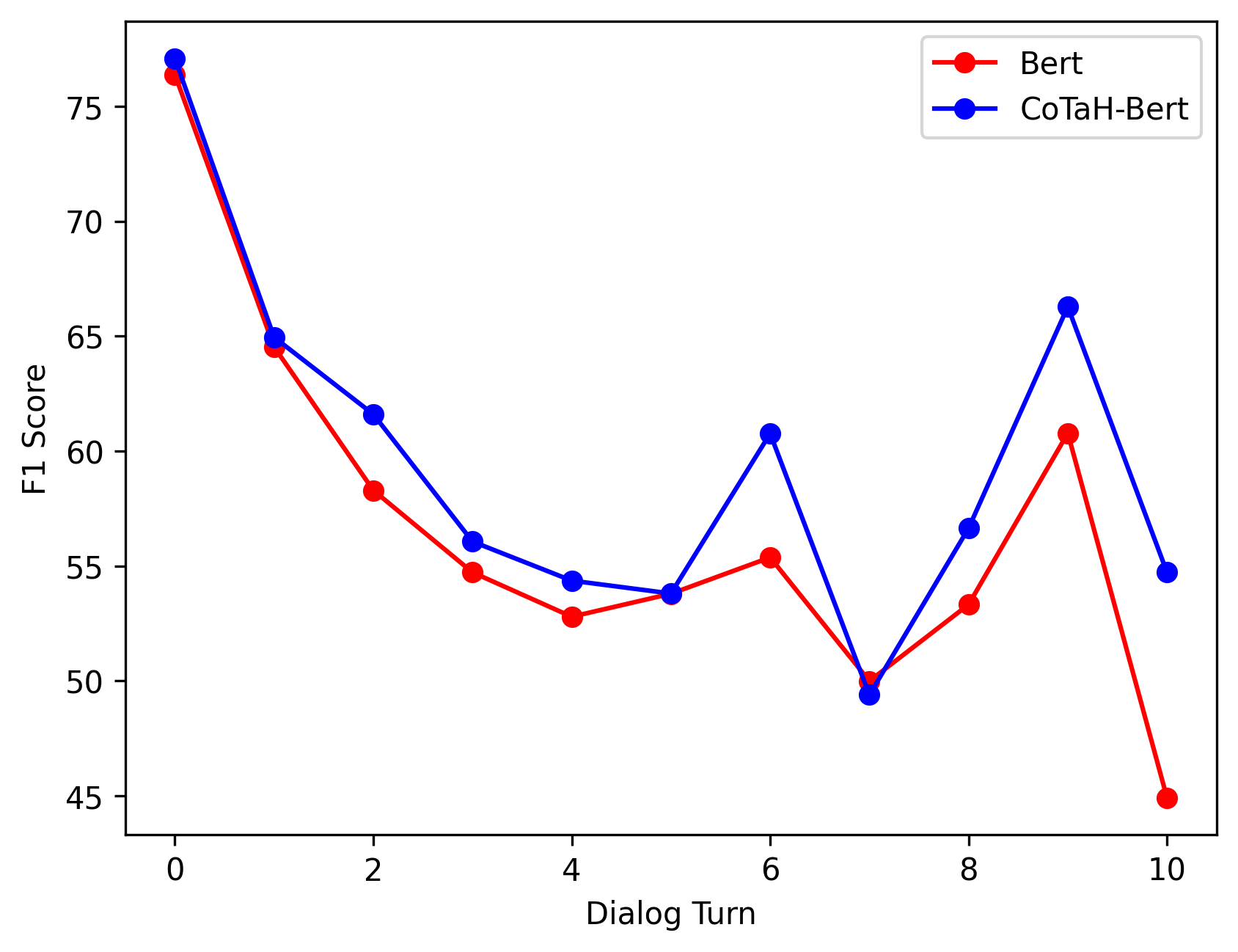}
 		\caption
 		{
             The F1 score of the test set dialog turns
 		}
 		\label{fig:test_turns}
 	\end{center}
 \end{figure}

In Figure \ref{fig:test_turns}, a comparison between the F1 scores of questions for each turn in Bert and CoTaH-Bert on the test set is presented. The score for the $k$-th turn represents the average F1 score for all questions in the $k$-th turn across all dialogs in the test set. Questions with a considerable amount of historical context are answered more effectively with our method. For $0 \leq k \leq 1$, the performances of both Bert and CoTaH-Bert are nearly equal, which is sensible as these questions contain little historical context, and thus, they have little irrelevant history. However, for most of $k > 1$ dialog turns, CoTaH-Bert outperforms Bert or it has on par performance with Bert. The performance upgrade is especially evident towards the end of dialogs, where questions contain significant historical context. This finding indicates the superiority of CoTaH-Bert over Bert in establishing greater robustness in answering these questions, by identifying and ignoring the irrelevant history turns. 

\begin{figure*}[t]
	\begin{center}
		\includegraphics[width=1.\textwidth]{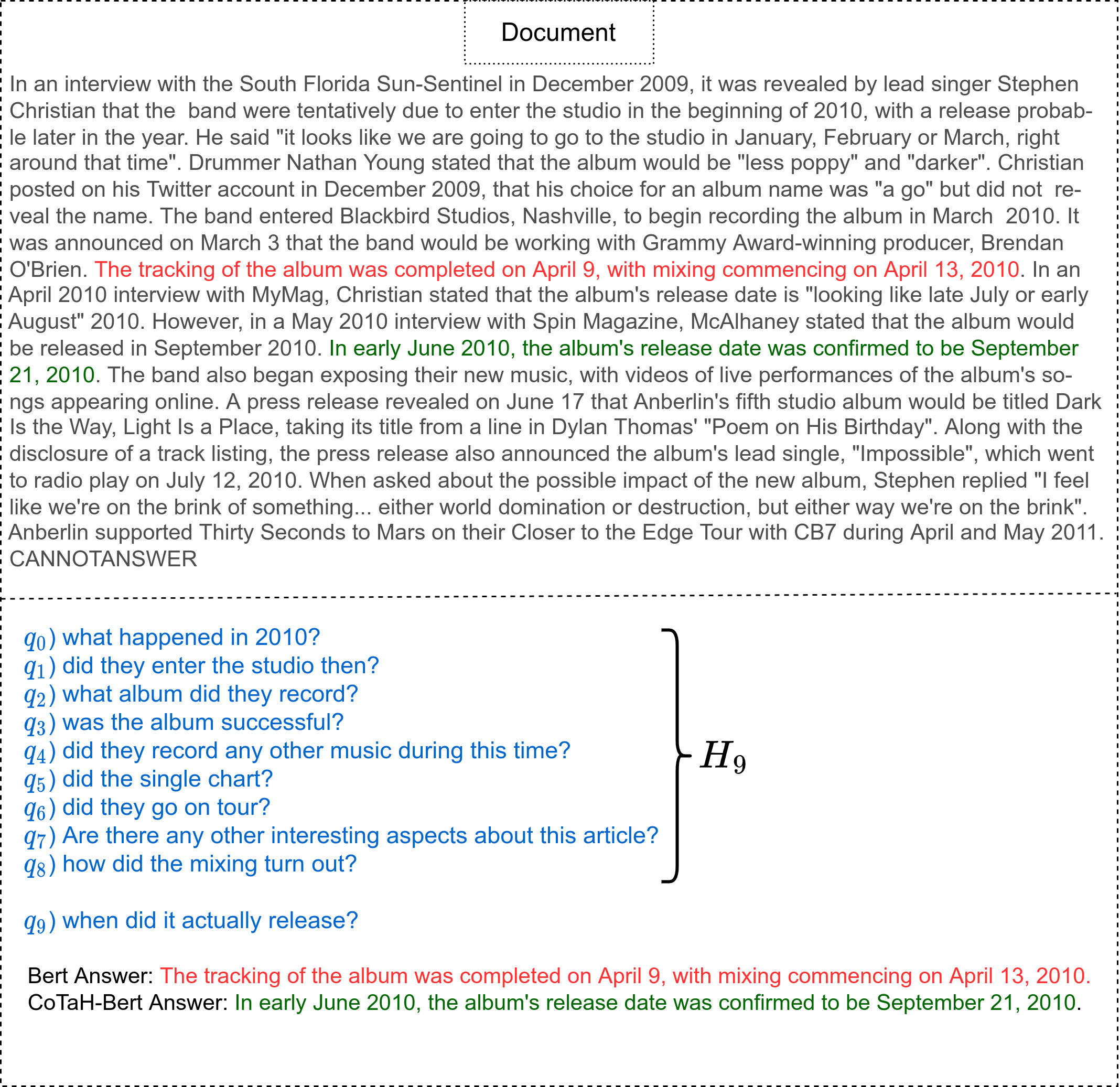}
		\caption
		{
A comparison between Bert and CoTaH-Bert extracted answers to a question, showing that CoTaH-Bert has been able to successfully ignore the irrelevant history by extracting the correct answer. However, the Bert model has been confused and returned a wrong answer.
		}
		\label{fig:sample}
	\end{center}
\end{figure*}

\subsection{Reproducibility} \label{rep}
The seed for all experiments, except the training of $CQG_\theta$, is 1000. All of the experiments to train the $QA_{\theta'}$ are conducted on a single RTX 3070 Ti with 8GB memory, on which each experiment takes approximately 6 hours. $CQG_\theta$ is trained on a single Tesla T4 from Google Colab. For each model, Bert or CoTaH-Bert, the hyperparameters are optimized on the dev set, and a final model will be trained on the train set with the optimized hyperparameters. Subsequently, a single result on the test set will be reported as depicted in Table \ref{table:table_comp}.

\subsection{Case Study} \label{case_study}
In Figure \ref{fig:sample}, a document sample with its corresponding dialog in the dev set is depicted. In the figure, ninth turn question, $q_9$, with its history, $H_9$, are shown. The answers of Bert and CoTaH-Bert to $q_9$ are compared, showing that CoTaH-Bert has been successful to answer this question with a full F1 score, while Bert has been unsuccessful. $q_9$ asks about the release date of the album stated in $q_2$. This is a suitable sample for our context, as there are significant irrelevant history turns between $q_9$ and $q_2$. We observe that CoTaH-Bert has been successful in identifying the relevant history by answering the question correctly. However, the Bert model has mistakenly reported another date which is wrong. As Bert has returned a span containing the word ``mixing'', it's possible that Bert has incorrectly identified the previous turn question, $q_8$, as relevant, and has returned a span by text matching encompassing the word ``mixing'', and containing merely some random dates.


\end{document}